\documentclass[a4paper, 10 pt, conference]{ieeeconf}
\IEEEoverridecommandlockouts
\usepackage{epsfig}
\usepackage{amsmath}
\usepackage{array}
\usepackage{amssymb}
\usepackage{cite}
\usepackage{booktabs}
\usepackage{multirow}
\usepackage{subcaption}
\usepackage{booktabs}
\usepackage{mathtools}
\usepackage{siunitx}
\usepackage{comment}
\usepackage{optidef}
\usepackage{threeparttable}
\usepackage{bm}
\usepackage{pifont}
\makeatletter
\newcommand{\subsubsubsection}{\@startsection{paragraph}{4}{\z@}%
  {1.0\Cvs \@plus.5\Cdp \@minus.2\Cdp}%
  {.1\Cvs \@plus.3\Cdp}%
  {\reset@font\sffamily\normalsize}
}
\makeatother
\newcommand{\figlab}[1]{\label{fig:#1}}
\newcommand{\figref}[1]{Fig.~\ref{fig:#1}} 

\newcommand{\forlab}[1]{\label{for:#1}}
\newcommand{\forref}[1]{Equation~(\ref{for:#1})} 
\newcommand{\seclab}[1]{\label{sec:#1}}
\newcommand{\secref}[1]{Section~\ref{sec:#1}} 

\newcommand{\algolab}[1]{\label{algorithm:#1}}
\newcommand{\algoref}[1]{Algorithm~\ref{algorithm:#1}} 

\newcommand{\etal}{\textit{et~al.}}
\newcommand{\ie}{\textit{i.e.}}
\newcommand{\eg}{\textit{e.g.}}
\usepackage{color}
\usepackage{soul}

\usepackage{amsmath}
\usepackage{algorithmicx}
\usepackage[ruled]{algorithm}
\usepackage[noend]{algpseudocode}

\begin{document}
\title{Hierarchical Planning and Scheduling for Reconfigurable Multi-Robot Disassembly Systems under Structural Constraints}

\author{Takuya Kiyokawa$^{1}$, Tomoki Ishikura$^{2}$, Shingo Hamada$^{2}$, Genichiro Matsuda$^{2}$, and Kensuke Harada$^{1}$%
\thanks{$^{1}$Department of Systems Innovation, Graduate School of Engineering Science, The University of Osaka, Toyonaka, Osaka, Japan.}%
\thanks{$^{2}$Manufacturing Innovation Division, Panasonic Holdings Corporation, 2-7 Matsuba-cho, Kadoma, Osaka, Japan.}
}

\maketitle

\begin{abstract}
This study presents a system integration approach for planning schedules, sequences, tasks, and motions for reconfigurable robots to automatically disassemble constrained structures in a non-destructive manner. Such systems must adapt their configuration and coordination to the target structure, but the large and complex search space makes them prone to local optima. To address this, we integrate multiple robot arms equipped with different types of tools, together with a rotary stage, into a reconfigurable setup. This flexible system is based on a hierarchical optimization method that generates plans meeting multiple preferred conditions under mandatory requirements within a realistic timeframe. The approach employs two many-objective genetic algorithms for sequence and task planning with motion evaluations, followed by constraint programming for scheduling. Because sequence planning has a much larger search space, we introduce a chromosome initialization method tailored to constrained structures to mitigate the risk of local optima. Simulation results demonstrate that the proposed method effectively solves complex problems in reconfigurable robotic disassembly.
\end{abstract}

\IEEEpeerreviewmaketitle

\section{Introduction}
In robotic disassembly of structurally constrained objects, the system must access the target from multiple directions and plan disassembly sequences, task allocations, manipulation motions, and execution schedules that respect part relationships.
Although task- or motion-level approaches have been studied~\cite{Ebinger2018,Laili2022,Wang2022,Kiyokawa2024,Hansjosten2024}, integrated methods that jointly consider sequence, task, motion, and schedule under structural constraints remain limited.

\figref{cell}~(a) shows the simulated work cell. The planning problem requires generating a disassembly sequence, manipulation motions with tool changes (\figref{cell}~(b)), and task allocation (\ie, assigning robots to tasks) under structural constraints, followed by a schedule that coordinates reorientation and multi-robot operations.

\begin{figure}[tb]
    \centering
    \begin{minipage}[tb]{\linewidth}
        \centering
        \includegraphics[width=\linewidth]{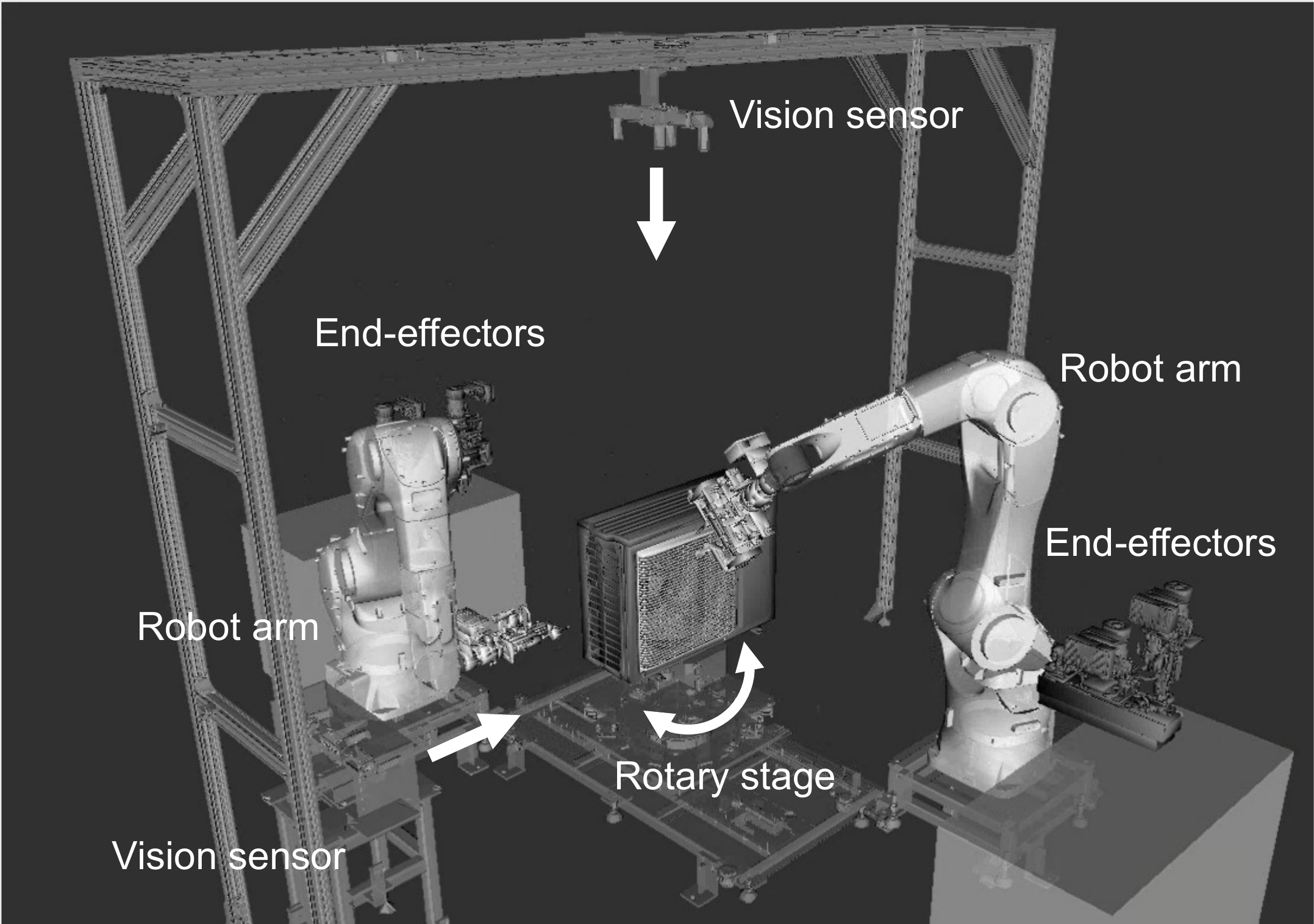}
        \subcaption{Simulated work cell}
    \end{minipage}
    \begin{minipage}[tb]{\linewidth}
        \centering
        \includegraphics[width=\linewidth]{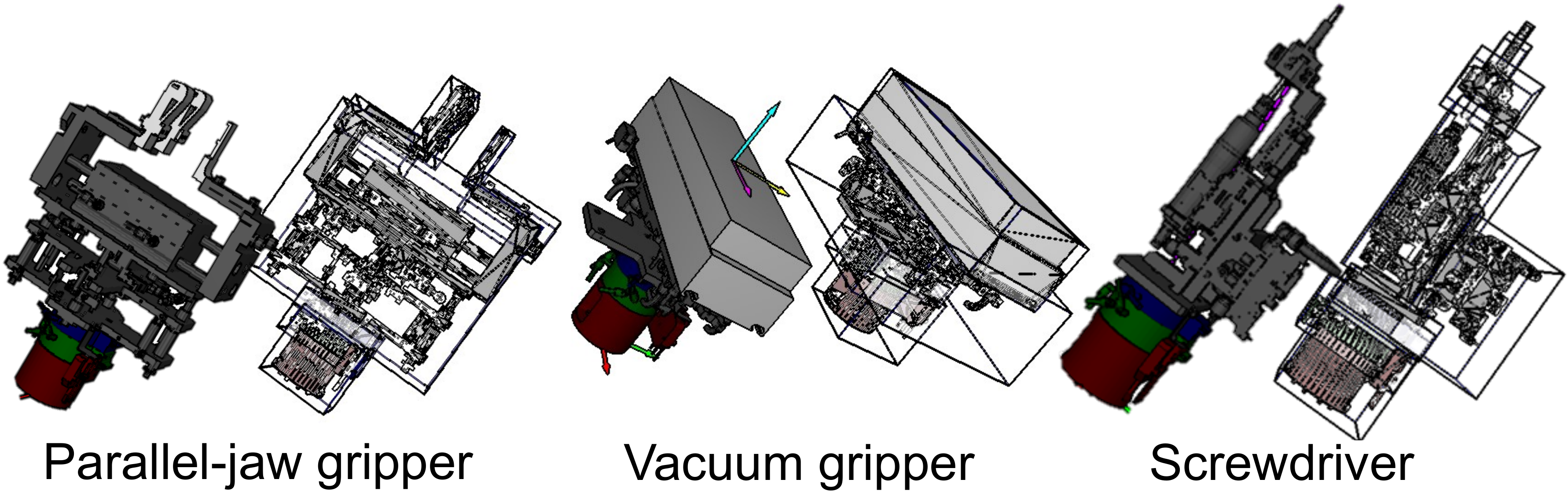}
        \subcaption{Tools}
    \end{minipage}
    \caption{Assumed robotic work cell for disassembly}
    \figlab{cell}
\end{figure}
\begin{figure}[tb]
    \centering
    \includegraphics[width=\linewidth]{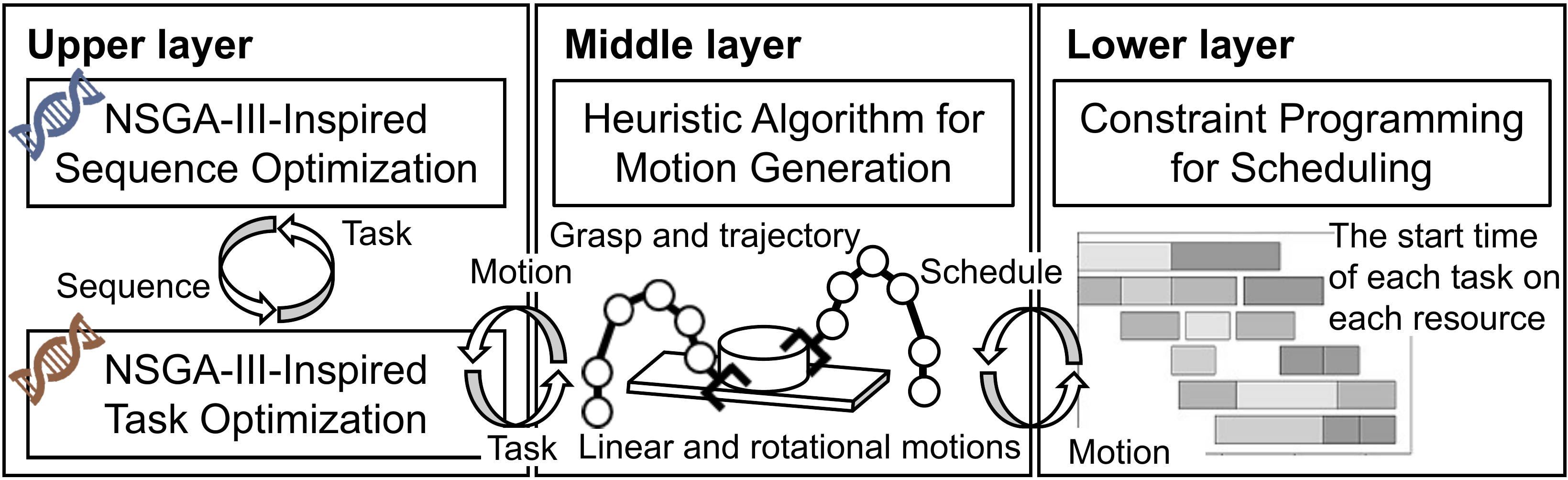}
    \caption{Overview of the hierarchical framework for generating disassembly sequence, task, motion, and scheduling}
    \figlab{ho}
\end{figure}
\figref{ho} shows the proposed hierarchical framework for solving the integrated planning-and-scheduling problem by progressively reducing the search space.
The framework addresses robotic disassembly of constrained assemblies using a reconfigurable multi-robot system equipped with a rotary stage for object reorientation.

Given the NP-hardness and PSPACE-completeness of sequence and task–motion planning~\cite{Goldwasser1996,Brown2020}, the problem is tackled within a hierarchical optimization scheme. Sequence and task planning are solved with a many-objective genetic algorithm (MOGA), while schedule feasibility is evaluated by constraint programming (CP).

At the upper level, an NSGA-III–inspired MOGA optimizes disassembly sequences and task allocation. Building on Kiyokawa~\etal~\cite{Kiyokawa2024}, chromosome initialization is extended to encode structural constraints, improving feasible and stable sequence generation.
At the middle level, a heuristic algorithm plans grasp and trajectory motions of multiple arms, together with object repositioning and reorientation by the rotary stage.
At the lower level, CP schedules tasks on available resources, computing start times under precedence and blocking constraints.

Simulations show that the proposed framework enhances feasibility and accelerates convergence in long-horizon multi-robot disassembly with structural constraints.

\section{Related Works}
\subsection{Robotic Disassembly}
Robotic disassembly has been studied in remanufacturing and recycling to automate component removal from complex products.
For structurally complex objects, Zhang~\etal~\cite{Zhang2023} used a PDDL-based TAMP framework with visual and force feedback, and Diaz~\etal~\cite{Diaz2025} applied force-sensing for precise unscrewing.
These studies advance specific elements such as sequencing, motion planning, and task allocation, but integrated planning and scheduling under structural constraints and long time horizons remains limited.

\subsection{Integrated Planning and Scheduling}
Integrated planning and scheduling have been explored in robotic manipulation under TAMP and CP frameworks.
Lagriffoul~\etal~\cite{Lagriffoul2014} highlighted the importance of geometric constraints in TAMP. 
Kiyokawa~\etal~\cite{Kiyokawa2024case} tackled solely generating disassembly sequences and motion plans from 3D CAD models under structural constraints.
Behrens~\etal~\cite{Behrens2019} developed a CP-based model with Ordered Visiting Constraints for multi-robot task allocation and scheduling. Tian~\etal~\cite{Tian2024} applied physics-based search to generate disassembly sequences for large assemblies.
Few approaches, however, combine hierarchical multi-objective optimization with constraint-based planning and scheduling for long-horizon, constrained disassembly, as in our work.

\section{Problem Setting} \seclab{problem}
A 3D CAD model is provided, with each part defined as a separate component mapped to a disassembly task: grasp, suction, or screw removal, executed by a parallel gripper, suction gripper, or air driver.
For each part, feasible handling orientations for both arms are given relative to the nominal pose; if multiple orientations exist, the smallest rotation is chosen.

The disassembly order of $\eta$ parts is $\bm{O}=[O_1,\ldots,O_\eta]$, where $O_1$ is the last part removed. In optimization, the $i$-th individual at generation $g$ is $\bm{O}_{g,i}$, with $i=1,\ldots,N^{\mathrm{g}}$, $g=1,\ldots,G$. The initial and optimized sequences are $\bm{O}^{\mathrm{initial}}$ and $\bm{\hat{O}}$. The optimized task sequence, arm assignment, motion plan, and schedule are $\bm{\hat{T}}$, $\bm{\hat{A}}$, $\bm{\hat{M}}$, and $\bm{\hat{S}}$.

The scheduling problem is defined as the resource set $\mathcal{R}=\{\text{Arm1\_S1},\text{Arm2\_S1},\text{ExternalAxis\_S2},\text{Arm1\_S3},\text{Arm2\_S3}\}$.
Operations are grouped into three stage types mapped to $\mathcal{R}$: tool change (stage~1), pose adjustment on the external axis (stage~2), and disassembly (stage~3). Precedence among disassembly jobs is induced by the sequence $\bm{\hat{O}}$.
Given the planned task set $\bm{\hat{T}}$, arm assignment $\bm{\hat{A}}$, and motion estimates $\bm{\hat{M}}$, each task $k\in\bm{\hat{T}}$ has a duration $d_k$ derived from $\bm{\hat{M}}$ and is assigned a resource $r_k\in\mathcal{R}$.
The schedule is expressed as $\bm{\hat{S}}=\{s_k\}$, where $s_k$ denotes the start time of task $k$.

This representation $(\bm{O},\bm{T},\bm{A},\bm{M},\bm{S})$ integrates sequence, task, and motion planning with resource-aware scheduling, enabling coordinated multi-robot disassembly under structural constraints.

\section{Methods}
\figref{ho} outlines a three-level hierarchy. At the upper level, sequence planning is posed as constrained optimization (\secref{sequence}) following \cite{Kiyokawa2024} and augmented by the proposed chromosome initialization (\secref{initialization}); this stage also optimizes arm assignment (\secref{task}), yielding $\bm{\hat{O}}$, $\bm{\hat{T}}$, and $\bm{\hat{A}}$. The middle level plans grasps and trajectories together with rotary stage reorientation to produce the motion plan $\bm{\hat{M}}$ (\secref{motion}). At the lower level, CP computes the schedule $\bm{\hat{S}}$ from task durations $d_k$ extracted from $\bm{\hat{M}}$ and the constraints defined in (\secref{problem}).

\subsection{Sequence Optimization} \seclab{sequence}
Following the structural analysis and matrix generation in~\cite{Kiyokawa2024}, feasibility and stability are evaluated from the generated matrices, and only sequences in $\mathcal{X}^{\mathrm{F}}\cap\mathcal{X}^{\mathrm{S}}$ are considered for optimization. Each sequence is evaluated by four objectives: constraint transition difficulty $f^{\mathrm{difficulty}}(\bm{O})$, sequence efficiency $f^{\mathrm{efficiency}}(\bm{O})$, priority-based ordering $f^{\mathrm{priority}}(\bm{O})$, and task allocation efficiency $f^{\mathrm{allocation}}(\bm{O},\bm{\hat{A}})$.
The first three objectives follow the definitions in~\cite{Kiyokawa2024}, while the last objective is computed as the summation of three elements defined in~\forref{task}.
Optimization is then performed using an MOGA to minimize these objectives.
\begin{customopti!}|s|
  {min}
  {\bm{O}}
  {\begin{bmatrix}
    f^{\mathrm{difficulty}}(\bm{O})\\
    f^{\mathrm{efficiency}}(\bm{O})\\
    f^{\mathrm{priority}}(\bm{O})\\
    f^{\mathrm{allocation}}(\bm{O},\bm{\hat{A}})
  \end{bmatrix}}
  {\forlab{seq}}
  {}
  \addConstraint{\bm{O}}{= [O_1,\ldots,O_\eta]}
  \addConstraint{O_j}{\in [1,\ldots,\eta],\; j\in\{1,\ldots,\eta\}}
  \addConstraint{\bm{O}}{\in \mathcal{X}^{\mathrm{F}} \cap \mathcal{X}^{\mathrm{S}}}.
\end{customopti!}

Initial solutions are generated by a contact--connection--constraint (CCC) graph-based chromosome initialization that promotes feasibility and stability. The search adopts NSGA-III~\cite{NSGAIII} for many-objective sequence generation setting, using non-dominated sorting and reference-direction niching; chromosomes encode permutations and genetic operators follow sequence-optimization practice~\cite{Tariki2021}. The loop of evaluation, selection, and variation repeats until the termination condition is met, after which the final solution set is returned.

\begin{figure}[tb]
    \centering
    \begin{minipage}[tb]{0.365\linewidth}
        \centering
        \includegraphics[width=\linewidth]{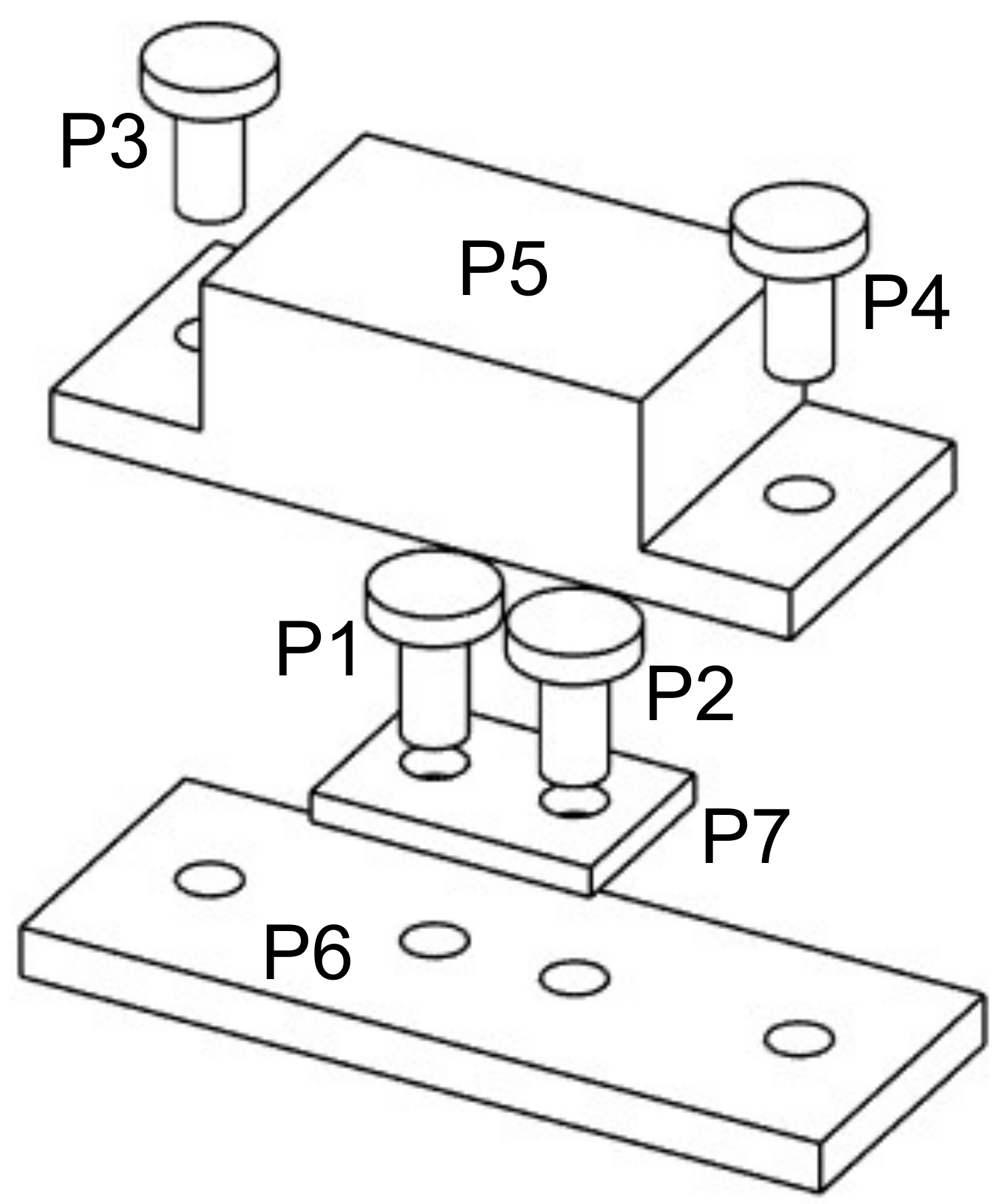}
        \subcaption{Constrained structure}
    \end{minipage}
    \begin{minipage}[tb]{0.62\linewidth}
        \centering
        \includegraphics[width=\linewidth]{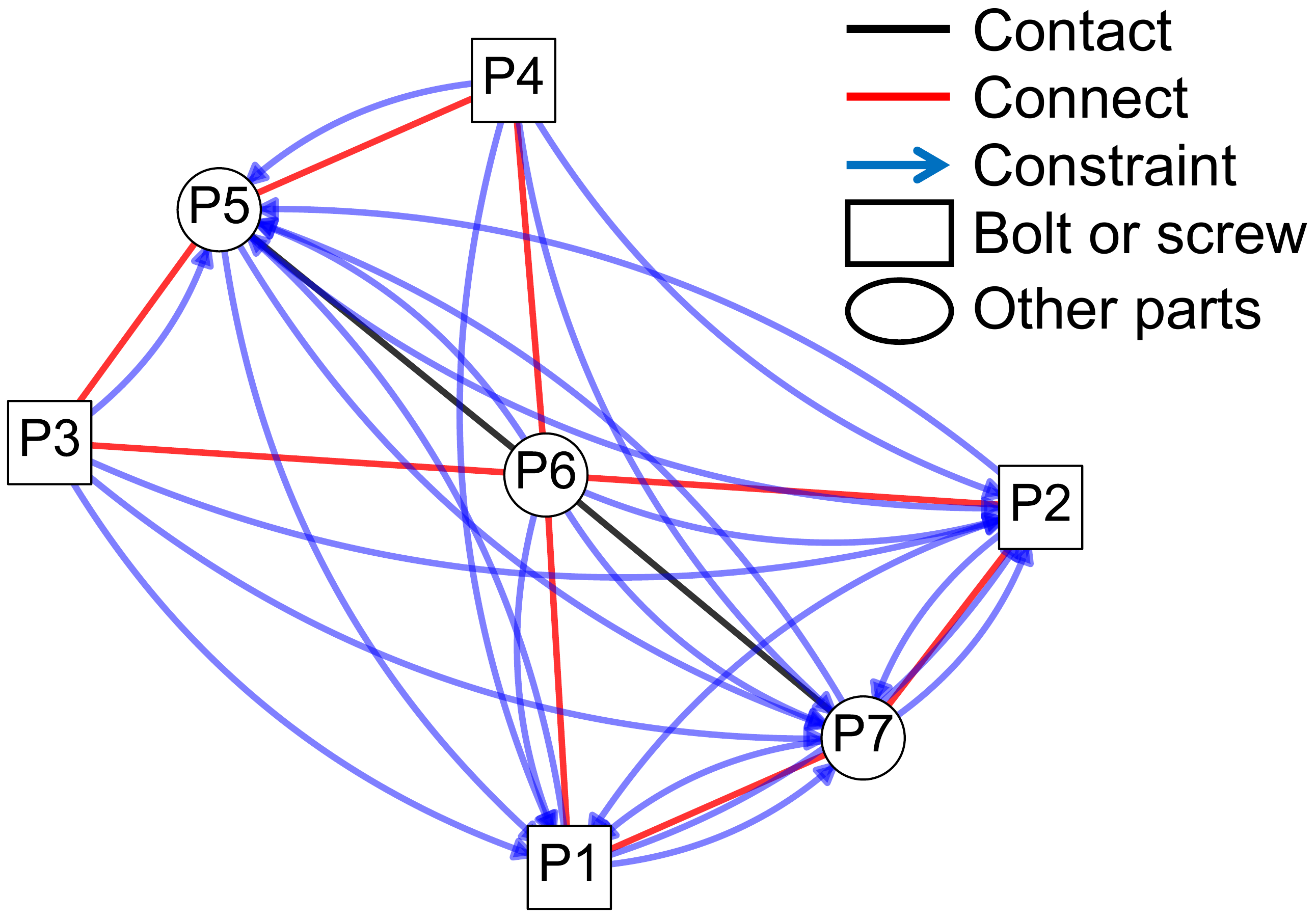}
        \subcaption{CCC graph}
    \end{minipage}
    \caption{Graphical representation of constrained structure by contact, connection, and constraint (CCC) graph}
    \figlab{graph}
\end{figure}

\subsection{Initialization Using Parts Relation (CCC) Graph} \seclab{initialization}
A CCC graph is constructed from the parts relation matrices $\bm{X}$ and labels $\bm{I}$, yielding node set $\bm{N}$ and link set $\bm{L}$. Candidates are filtered by the order and motion feasibility check; fasteners are prioritized; non-fastener parts are admitted only after their fasteners are removed; the selected node is prepended so that $\bm{O}^{\mathrm{initial}}$ is built from the last-to-be-removed side. Among feasible candidates, the node farthest from the base $b$ (shortest-path in the undirected CCC graph) is preferred to avoid isolating substructures.

\figref{graph}~(a) and (b) provide a minimal example: P1–P4 are screws; P5 and P7 have through-holes; P6 has threaded holes; P5 and P7 are fixed to P6 via \{P3,P4\} and \{P1,P2\}. Contacts (\eg, P6--P5, P6--P7) and connections (\eg, P1--P7, P1--P6, \dots, P4--P6) are undirected; constraints such as P5$\to$P1, P5$\to$P2, P5$\to$P7 are directed. This encoding enforces fastener precedence and exposes feasible removals.

\alglanguage{pseudocode}
\begin{algorithm}[t]
\algrenewcommand\algorithmicindent{0.8em}
\algrenewcommand\algorithmiccomment[1]{\hfill$\triangleright$\,#1}
\caption{CCC-Graph-Based Initialization} \algolab{prgbi}
\begin{algorithmic}[1]
\renewcommand{\algorithmicrequire}{\textbf{Input:}}
\renewcommand{\algorithmicensure}{\textbf{Output:}}
\Require Parts count $\eta$, labels $\bm{I}$, matrices $\bm{X}$, gene count $N^{\mathrm{g}}$
\Ensure Initial chromosomes $\{\bm{O}^{\mathrm{initial}}_l\}_{l=1}^{N^{\mathrm{g}}}$
\State $\bm{N},\bm{L} \leftarrow \textsc{BuildGraph}(\bm{I},\bm{X})$;\; $b \leftarrow \textsc{BaseNode}(\bm{I})$
\For{$l \leftarrow 1$ \textbf{to} $N^{\mathrm{g}}$}
  \State $\bm{O}^{\mathrm{initial}}_l \leftarrow [\,]$;\; $\bm{N}^t,\bm{L}^t \leftarrow \bm{N},\bm{L}$
  \While{$\bm{N}^t \setminus \{b\} \neq \varnothing$}
    \State $\bm{D} \leftarrow \textsc{DistanceFromBase}(\bm{N}^t,\bm{L}^t,b)$
    \State $\mathcal{R} \leftarrow \bm{N}^t \setminus \bigl(\{b\}\cup \textsc{ReachableNodes}(\bm{D})\bigr)$
    \State $\mathcal{L} \leftarrow \textsc{Layers}(\bm{D},\mathcal{R})$
    \State $\mathcal{U} \leftarrow \textsc{CnstraintFree}(\bm{N}^t,\bm{L}^t,\bm{X})$
    \State $\textit{Placed} \leftarrow \textbf{false}$
    \For{\textbf{each} layer $L \in \mathcal{L}$}
      \State $\mathcal{M} \leftarrow \textsc{CollisionFree}(L,\bm{L}^t,\bm{X})$
      \State $\mathcal{A} \leftarrow \mathcal{M} \cap \mathcal{U}$
      \If{$\mathcal{A} \neq \varnothing$}
        \State $\mathcal{F} \leftarrow \textsc{Fasteners}(\mathcal{A})$
        \State $\mathcal{C} \leftarrow \textsc{ConnectFree}(\mathcal{F},\mathcal{A},\bm{N}^t,\bm{L}^t)$
        \If{$\mathcal{C} \neq \varnothing$}
          \State $g \leftarrow \textsc{RandomPick}(\mathcal{C})$
          \If{\textsc{Fasners}$(g,\bm{L}^t)$}
            \State $g \leftarrow \textsc{RandomPick}(\textsc{Fasteners}(g,\bm{L}^t))$
          \EndIf
          \State \textsc{Prepend}$(\bm{O}^{\mathrm{initial}}_l,\textsc{PartID}(g))$
          \State \textsc{Remove}$(g,\bm{N}^t,\bm{L}^t)$
          \State $\textit{Placed} \leftarrow \textbf{true}$;\; \textbf{break}
        \EndIf
      \EndIf
    \EndFor
    \If{\textit{Placed} = \textbf{false}}
      \State \textbf{break}
    \EndIf
  \EndWhile
\EndFor
\end{algorithmic}
\end{algorithm}

\algoref{prgbi} constructs initial chromosomes from the structural analysis outputs \((\eta,\bm{I},\bm{X})\) and the gene count \(N^{\mathrm{g}}\) predefined.
A CCC graph with nodes \(\bm{N}\) and links \(\bm{L}\) is built from \(\bm{I}\) and \(\bm{X}\), and the base node \(b\) is identified; in \figref{graph}, the base node \(b\) is P6.
For each chromosome \(l=1,\ldots,N^{\mathrm{g}}\), a working graph \((\bm{N}^t,\bm{L}^t)\) is reset and the sequence \(\bm{O}^{\mathrm{initial}}_l\) is grown last-to-first.

At each step, distance layers from the base are computed by \textsc{DistanceFromBase}, and nodes not reachable from the base are grouped into \(\mathcal{R}\).
The ordered list of candidate layers \(\mathcal{L}\) is then constructed by placing \(\mathcal{R}\) first, followed by the distance layers arranged from farthest to nearest.
The set \(\mathcal{U}\) of nodes with constraint-edge in-degree zero is recomputed on the current CCC subgraph.
Within each layer \(L\in\mathcal{L}\), we evaluate \(\mathcal{M}=\textsc{CollisionFree}(L)\) by direction-wise interference checks and restrict it to \(\mathcal{A}=\mathcal{M}\cap\mathcal{U}\).
If \(\mathcal{A}\) is nonempty, fasteners are preferred; otherwise connector-free regular parts are considered.
\textsc{PickFar} then selects one node uniformly at random from the first nonempty layer encountered in the far-to-near order.
If the chosen node is a non-fastener with direct fasteners still attached, one of its direct fasteners is detached instead.
Here, \textsc{HasFasteners}\((n,\bm{L}^t)\) holds if there exists a fastener \(f\) such that \((n,f)\) is a connection edge in \(\bm{L}^t\), and \textsc{Fasteners}\((n,\bm{L}^t)\) denotes the set of such incident fasteners.
The selected node is prepended to \(\bm{O}^{\mathrm{initial}}_l\) and removed from the working graph.
If no admissible node is found in any layer, the loop terminates.
Repeating this procedure for each \(l\) yields the population \(\{\bm{O}^{\mathrm{initial}}_l\}_{l=1}^{N^{\mathrm{g}}}\), biased toward connector-cleared and constraint-admissible choices and toward nodes far from \(b\), which helps avoid isolating substructures and promotes feasibility and stability.

\subsection{Task Optimization} \seclab{task}
Given a feasible disassembly sequence $\bm{O}$, the corresponding task order $\bm{\hat{T}}$ is fixed, while the arm assignment $\bm{A}$ is optimized using the NSGA-III-inspired MOGA. 
Each chromosome encodes the arm IDs assigned to the parts in $\bm{O}$. 
The objectives are designed to encourage effective parallelization and to penalize excessive tool changes and inter-part travel. 
The optimization problem is formulated as
\begin{customopti!}|s|
  {min}
  {\bm{A}}
  {\begin{bmatrix}
    f^{\mathrm{parallel}}(\bm{O},\bm{A})\\
    f^{\mathrm{change}}(\bm{O},\bm{A})\\
    f^{\mathrm{distance}}(\bm{O},\bm{A})
  \end{bmatrix}}
  {\forlab{task}}
  {}
  \addConstraint{\bm{A}}{= [A_1,\ldots,A_\eta] \forlab{task_const1}}
  \addConstraint{A_i}{\in [1,2],\; i\in\{1,\ldots,\eta\} \forlab{task_const2}}
  \addConstraint{\bm{O}}{= [O_1,\ldots,O_\eta] \forlab{task_const3}}
  \addConstraint{O_j}{\in [1,\ldots,\eta],\; j\in\{1,\ldots,\eta\} \forlab{task_const4}}
  \addConstraint{\bm{O}}{\in \mathcal{X}^{\mathrm{F}} \cap \mathcal{X}^{\mathrm{S}} \forlab{task_const5}}.
\end{customopti!}

Here, the objective vector consists of three efficiency terms: 
$f^{\mathrm{parallel}}$ (effective parallelized tasks), 
$f^{\mathrm{change}}$ (tool-change count), and 
$f^{\mathrm{distance}}$ (cumulative inter-part travel per arm). 
For the solutions in the admissible set 
$\mathcal{X}^{\mathrm{A}} \coloneqq \mathcal{X}^{\mathrm{F}} \cap \mathcal{X}^{\mathrm{S}}$, 
the normalized objectives $f^{\mathrm{parallel}}$, $f^{\mathrm{change}}$, and $f^{\mathrm{distance}}$ 
are defined as $C^{\mathrm{parallel}}/\eta$, $C^{\mathrm{change}}/(\eta-1)$, and $S^{\mathrm{distance}}/(\eta \, f^{\mathrm{distance}}_{\max})$, respectively, where $C^{\mathrm{parallel}}$, $C^{\mathrm{change}}$, and $S^{\mathrm{distance}}$ denote the raw counts or sums, 
and $f^{\mathrm{distance}}_{\max}$ is the maximum distance between the centers of mass of any two parts.

\subsection{Generating Motions} \seclab{motion}
No GA is used at this stage. Given $(\bm{\hat{T}},\bm{\hat{A}})$, a grasp plan is generated for the assigned tool with GraspGen~\cite{Murali2025}, then a system motion $\bm{\hat{M}}$ is synthesized by a fast heuristic: the rotary stage is oriented toward the assigned arm to reduce approach distance, and per-grasp trajectories are planned by RRT-Connect. If no feasible path exists, a small translational offset of the rotary stage is sampled and planning is retried; among feasible candidates, the shortest path is selected. 

For screw, bolt, and nut removal, an additional constraint is imposed that the rotary stage must be rotated such that the fastener head directly faces one of the recognition cameras mounted above or on the side, in order to enable reliable visual detection before disassembly.
\figref{cell}~(a) shows the camera position and orientations.

\subsection{Scheduling via Constraint Programming}\seclab{schedule}
Constraint programming (CP) is applied to determine the start times $\bm{\hat{S}}=\{s_k\}$ of the tasks.
Each task $k\in\bm{\hat{T}}$ has a processing time $d_k$ obtained from $(\bm{\hat{T}},\bm{\hat{A}},\bm{\hat{M}})$, including the fixed $2\,\mathrm{s}$ duration for pose adjustment when the selected rotary stage orientation differs from $0^\circ$.
To refer to the task of stage $p$ of job $J_i$, we denote it by $\kappa(i,p)\in\bm{\hat{T}}$ whenever that stage exists.

Intra-job precedence is enforced as
\begin{equation}
  s_{\kappa(i,p)} + d_{\kappa(i,p)} \le s_{\kappa(i,p+1)} \quad (p=1,2).
\end{equation}

Cross-stage exclusions are also imposed. Stage~2 and stage~3 are never executed in parallel, and stage~1 and stage~2 are also prohibited from overlapping. All stage~3 tasks are serialized according to the optimized disassembly order $\bm{\hat{O}}$. Stage~1 and stage~3 on the same arm are treated as mutually exclusive, requiring that one finishes before the next starts, while concurrency is permitted across arms when precedence allows.

The objective is makespan minimization,
\begin{equation}
  \min\ \max_{k\in\bm{\hat{T}}} (s_k+d_k),
\end{equation}
and among optimal schedules the earliest-start feasible one is selected.

\section{Experiments}
\subsection{Overview}
\figref{obj} shows the two CAD models used in our experiments, a condenser unit with 74 parts (Object~A) and a television with 133 parts (Object~B). For each object, the recovery target (motor (25th) for A, controller board (64th) for B) is highlighted in the center column; exploded views appear on the right. The proposed hierarchy, consisting of sequence, task, motion, and schedule, was applied end-to-end. The sequence and task layers used the NSGA-III–inspired MOGA, motion was generated heuristically using GrasgGen and RRT-Connect, and the final schedule was solved by the proposed CP. All runs were executed on an Intel Core i7-12700H.
\begin{figure}[tb]
    \begin{minipage}[tb]{\linewidth}
        \centering
        \includegraphics[width=\linewidth]{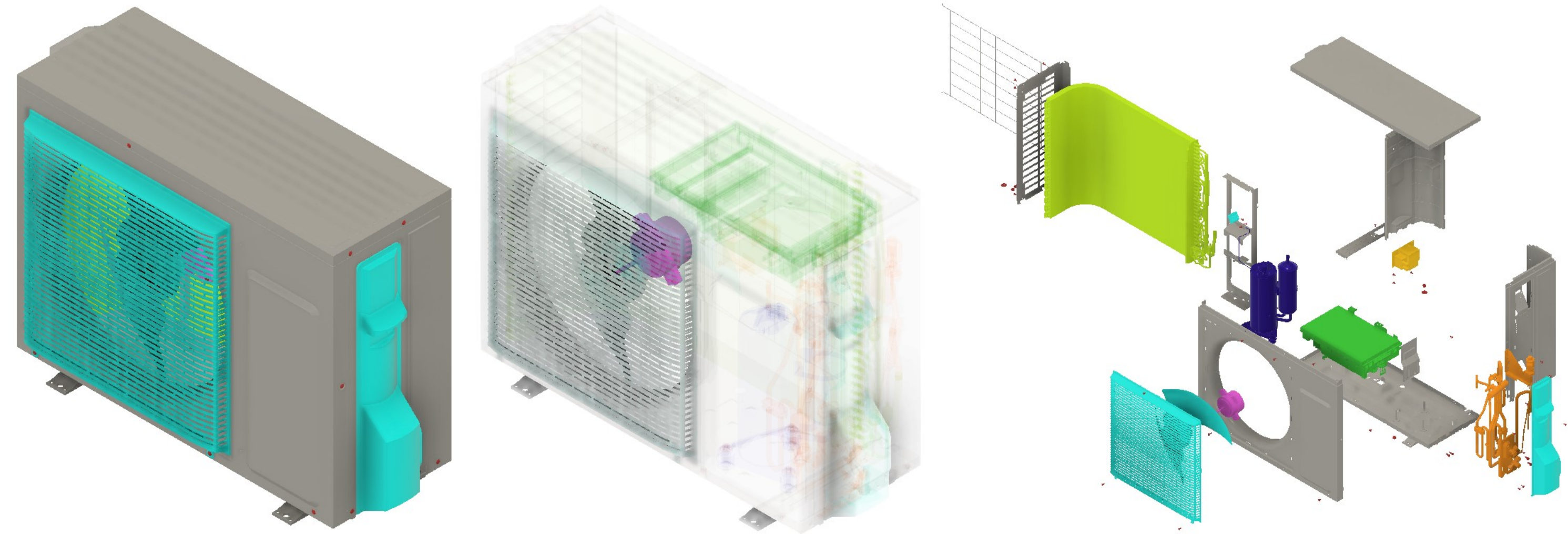}
        \subcaption{Condenser unit (Object A)}
    \end{minipage}
    \begin{minipage}[tb]{\linewidth}
        \centering
        \includegraphics[width=\linewidth]{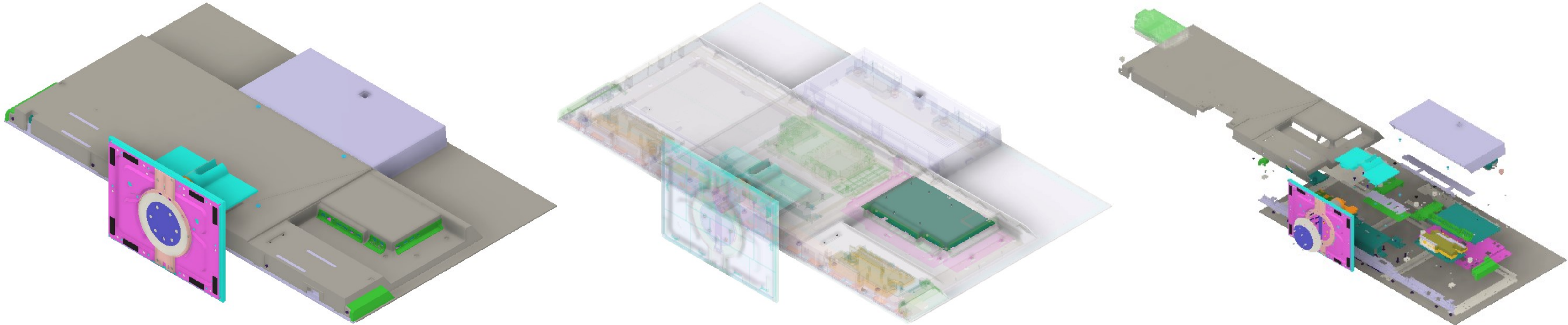}
        \subcaption{Television (Object B)}
    \end{minipage}
    \caption{3D models used for the experiments}
    \figlab{obj}
\end{figure}

\subsection{Results}
\begin{figure*}[tb]
    \begin{minipage}[tb]{\linewidth}
        \centering
        \includegraphics[width=\linewidth]{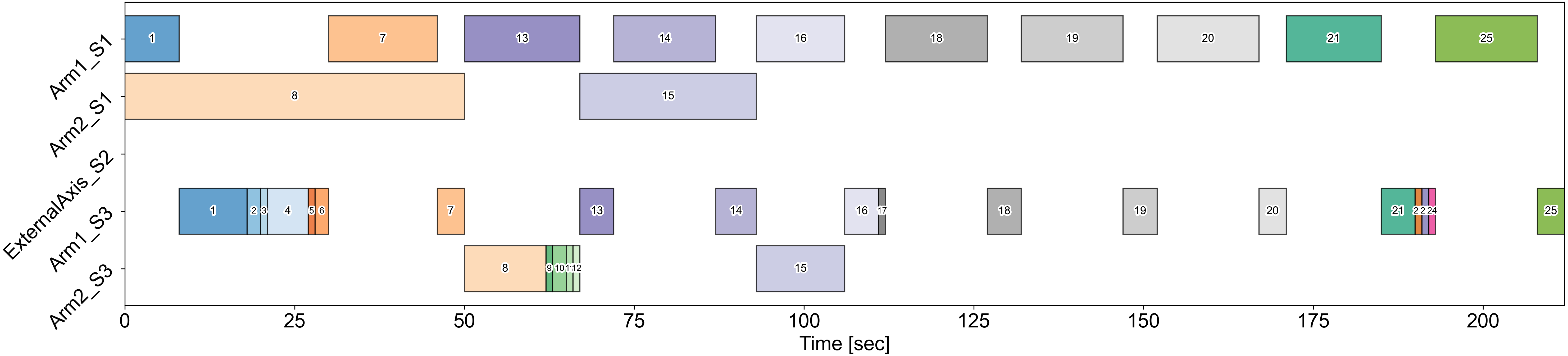}
        \subcaption{Condenser unit (Object A)}
    \end{minipage}
    \begin{minipage}[tb]{\linewidth}
        \centering
        \includegraphics[width=\linewidth]{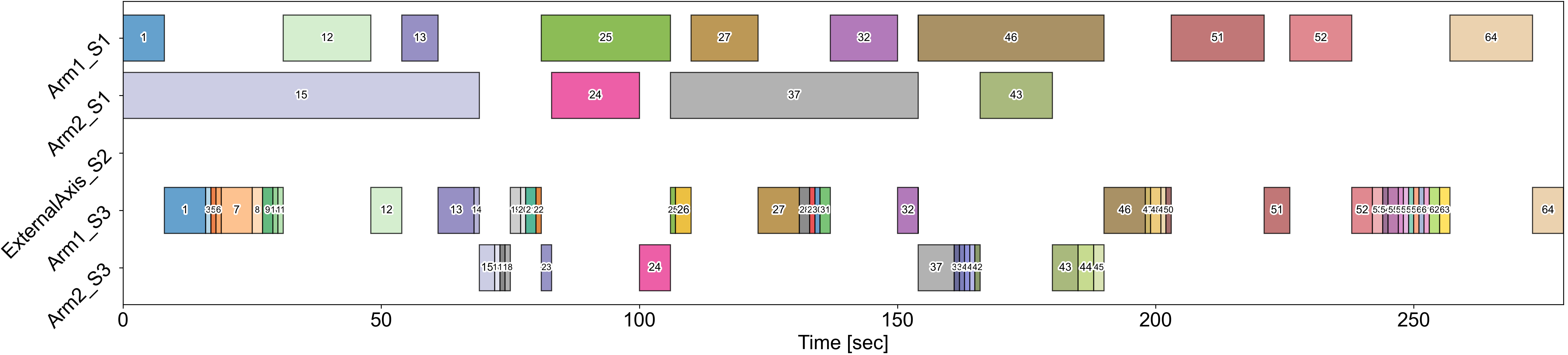}
        \subcaption{Television (Object B)}
    \end{minipage}
    \caption{Scheduling results without the external axis device}
    \figlab{schedule}
\end{figure*}
\begin{figure*}[tb]
    \begin{minipage}[tb]{\linewidth}
        \centering
        \includegraphics[width=\linewidth]{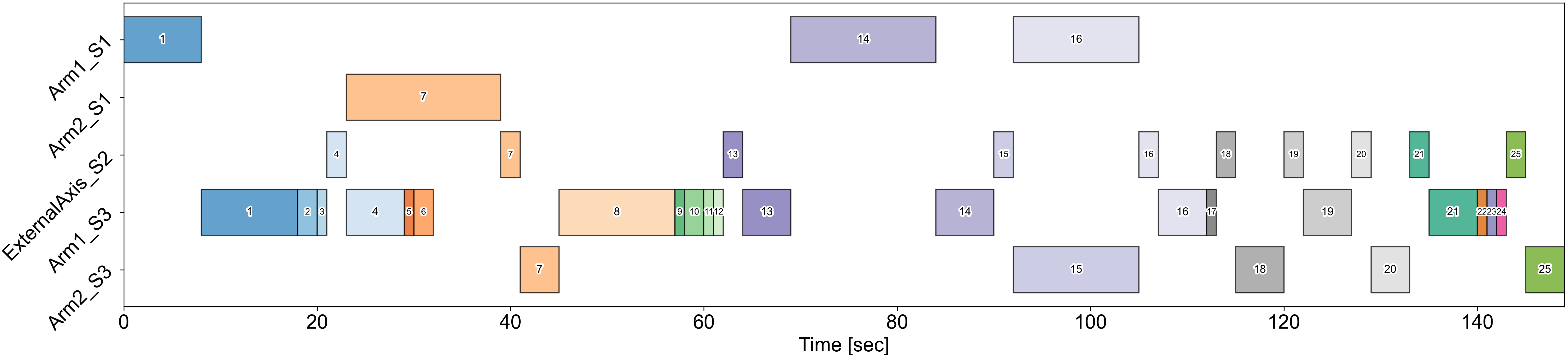}
        \subcaption{Condenser unit (Object A)}
    \end{minipage}
    \begin{minipage}[tb]{\linewidth}
        \centering
        \includegraphics[width=\linewidth]{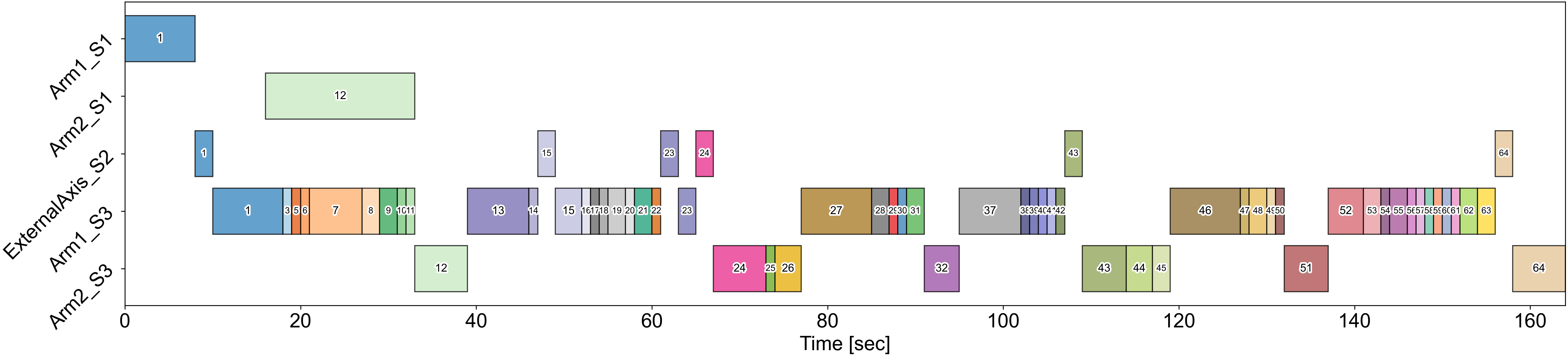}
        \subcaption{Television (Object B)}
    \end{minipage}
    \caption{Scheduling results with the external axis device}
    \figlab{schedule_wS2}
\end{figure*}
\figref{schedule} and \figref{schedule_wS2} show the computed schedules, with each row representing a resource (\text{Arm1}, \text{Arm2}, \text{ExternalAxis}) and each colored bar indicating a task interval. CP assigns independent tasks to different arms and inserts short reorientation tasks on the external axis only when required, thereby avoiding overlap with conflicting arm motions. In our simulations, the planned grasps, trajectories, and rotary stage motions were executed successfully, confirming the geometric consistency of $\bm{\hat{M}}$ with the upstream plans.

Quantitatively, without the external axis (\figref{schedule}), the total execution times were 216~s and 272~s for objects A and B. With the axis (\figref{schedule_wS2}), they were reduced to 146~s and 164~s. The number of tool changes decreased from 12 and 14 to 4 and 2, which mitigates wear in precision detachment operations.
These results demonstrate the effectiveness of the proposed scheduling approach. 
CP returned feasible earliest-start schedules for more than 95\% of valid inputs and reduced makespan by at least 32\% and number of toll changes by at least 67\% compared with the comparison method without using the rotary stage.
In \figref{schedule}, this appears as alternating arm utilization and short, well-placed rotary stage reorientations that keep both arms busy while maintaining safety separation on the external axis.

For chromosome initialization, the proposed CCC-graph method produced available initial chromosomes in every run for both objects, indicating that 100\% of the solutions were available. In contrast, the earlier CC-graph initialization yielded only 22\% (A) and 46\% (B) available solutions under comparable conditions. With the proposed CCC-graph initialization, the many-objective score always decreased from the generation-0 value, with the average reduction exceeding 20\% across runs. These observations match the design intent of the initializer, which encodes contact, connection, and explicit constraint edges, applies a feasibility filter, and biases picks toward nodes far from the base, collectively suppressing isolated subgraphs and interference at the start of the search. Starting from these populations, sequence optimization produced a high fraction of available sequences, and task optimization subsequently improved all evaluation values. The motion layer found collision-free and IK-solvable trajectories for all constraint-satisfying sequences using RRT-Connect and small rotary stage offsets when necessary, and feasible grasps were also obtained without difficulty.

\section{Discussion}
\begin{figure}[tb]
    \centering
    \includegraphics[width=\linewidth]{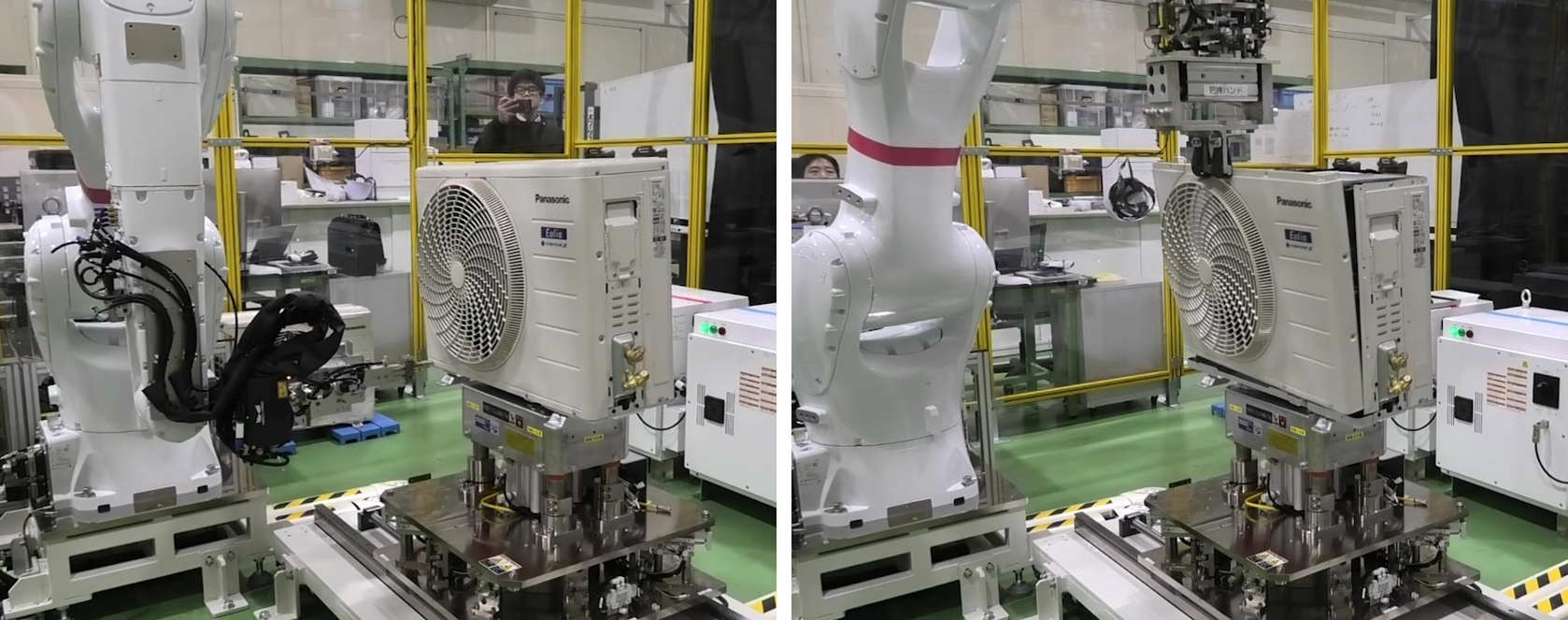}
    \caption{Real-world disassembly experiments}
    \figlab{real}
\end{figure}
The combination of CCC-graph initialization and CP-based scheduling demonstrates the benefit of treating structural constraints and execution resources within one pipeline.
The CCC graph enforces feasibility from the outset, preventing the search by the MOGA from wasting effort on unreachable or conflicting subproblems.
This is particularly valuable for large, multi-part products, where early pruning of infeasible sequences stabilizes optimization and improves convergence reliability.
The CP layer complements this by translating high-level plans into resource-feasible schedules that exploit multi-arm parallelism while respecting rotary stage and axis constraints.
Explicitly modeling resource contention prevents collisions and idle-time spikes that can otherwise arise even in geometrically valid plans.

\figref{real} shows the real-world disassembly experiments.
The execution of disassembly motions (\eg, removals of the screw and front cover) integrates a vision-based recognition method for the screw and an approach-pose adjustment procedure to ensure accurate tool engagement.
Our integrated system successfully removed them in our preliminary experiments, assuming our resulted plan and schedule.

In the present experiments, the scheduling model prohibited the simultaneous execution of separate S3 disassembly operations by multiple robots.
The translational device under the target object was also assumed to be unavailable.
These constraints were introduced to avoid potential robot–robot interference, to account for the lack of a vision system capable of recognizing multiple fasteners in parallel, and to prevent excessive loading that could damage the product.
Future work will address these limitations by modifying the scheduling constraints, developing a vision system capable of parallel recognition, and performing stability analysis based on physics simulation.
These would further close the gap between static plan generation and adaptive execution, enhancing safety and efficiency for reconfigurable disassembly.

\section{Conclusion}
This study targets long-horizon disassembly of structurally constrained assemblies, requiring joint optimization of sequence, task, motion, and schedule.
A chromosome initialization method based on a contact–connection–constraint relationship graph improves the generation of feasible and stable sequences.
A hierarchical optimization framework integrates a many-objective genetic algorithm for sequence and task planning with constraint programming for scheduling, enabling coordinated disassembly with multiple robots.

Simulation results show superior feasibility and faster convergence compared to conventional approaches.
Future work will extend the framework to other disassembly scenarios and evaluate its performance on real robotic systems.

\section*{Acknoledgement}
This work was supported by the New Energy and Industrial Technology Development Organization (NEDO) project JPNP23002. 

\bibliographystyle{IEEEtran}
\footnotesize
\bibliography{reference}

\end{document}